\documentclass[10pt,twocolumn,letterpaper]{article}

\usepackage[pagenumbers]{cvpr} 

\usepackage[dvipsnames]{xcolor}

\newcommand{\mbb}[1]{\ensuremath{\mathbb{#1}}}

\newcommand{\ve}[1]{\ensuremath{\mathbf{#1}}} 

\def\R{\mbb R}

\newcommand{\comment}[1]{}

\usepackage{multirow}
\usepackage{amssymb}
\usepackage{amsmath}
\usepackage{booktabs}
\usepackage{mathtools}
\usepackage{color}
\usepackage{pifont}
\newcommand{\xmark}{\ding{55}}%

\newcommand\blfootnote[1]{%
  \begingroup
  \renewcommand\thefootnote{}\footnote{#1}%
  \addtocounter{footnote}{-1}%
  \endgroup
}

\definecolor{cvprblue}{rgb}{0.21,0.49,0.74}
\usepackage[pagebackref,breaklinks,colorlinks,citecolor=cvprblue]{hyperref}
\usepackage[capitalize]{cleveref}
\def\paperID{22} 
\def\confName{CV4Animals}
\def\confYear{2024}

\title{Addressing the Elephant in the Room: Robust Animal Re-Identification with Unsupervised Part-Based Feature Alignment}

\author{Yingxue Yu \qquad
    Vidit Vidit \qquad
    Andrey Davydov \qquad
    Martin Engilberge \qquad
    Pascal Fua \\
    EPFL \qquad
    \\
    {\tt\small {\{firstname.lastname\}}@epfl.ch} \qquad
}

\begin{document}
\maketitle
\begin{abstract}
Animal Re-ID is crucial for wildlife conservation, yet it faces unique challenges compared to person Re-ID. \emph{First}, the scarcity and lack of diversity in datasets lead to background-biased models. \emph{Second}, animal Re-ID depends on subtle, species-specific cues, further complicated by variations in pose, background, and lighting. This study addresses background biases by proposing a method to systematically remove backgrounds in both training and evaluation phases. And unlike prior works that depend on pose annotations, our approach utilizes an unsupervised technique for feature alignment across body parts and pose variations, enhancing practicality. Our method achieves superior results on three key animal Re-ID datasets: ATRW, YakReID-103, and ELPephants.


\end{abstract}

\section{Introduction}
\label{sec:intro}

While Person Re-identification (Re-ID) has seen considerable advancements \cite{ye_deep_2022,LIU2020102018,wu_deep_2019,he_transreid_2021,he2022transfg,zhang2018shufflenet,he2018deep,chen2020orientation,wang2020smoothing,wang2018learning,sun_beyond_2018,khorramshahi2020devil,radford2021learning,li2023clip}, animal Re-ID~\cite{korschens_elpephants_2019,li_atrw_2020,zhang_yakreid-103_2021} remains underexplored despite its significance in both industrial and conservation contexts. Traditional animal identification methods are labor-intensive and impractical for large populations, highlighting the need for improved automated Re-ID techniques \cite{Martin2007ImportanceOW}.

\begin{figure}
    \centering
    \includegraphics[width=0.9\linewidth]{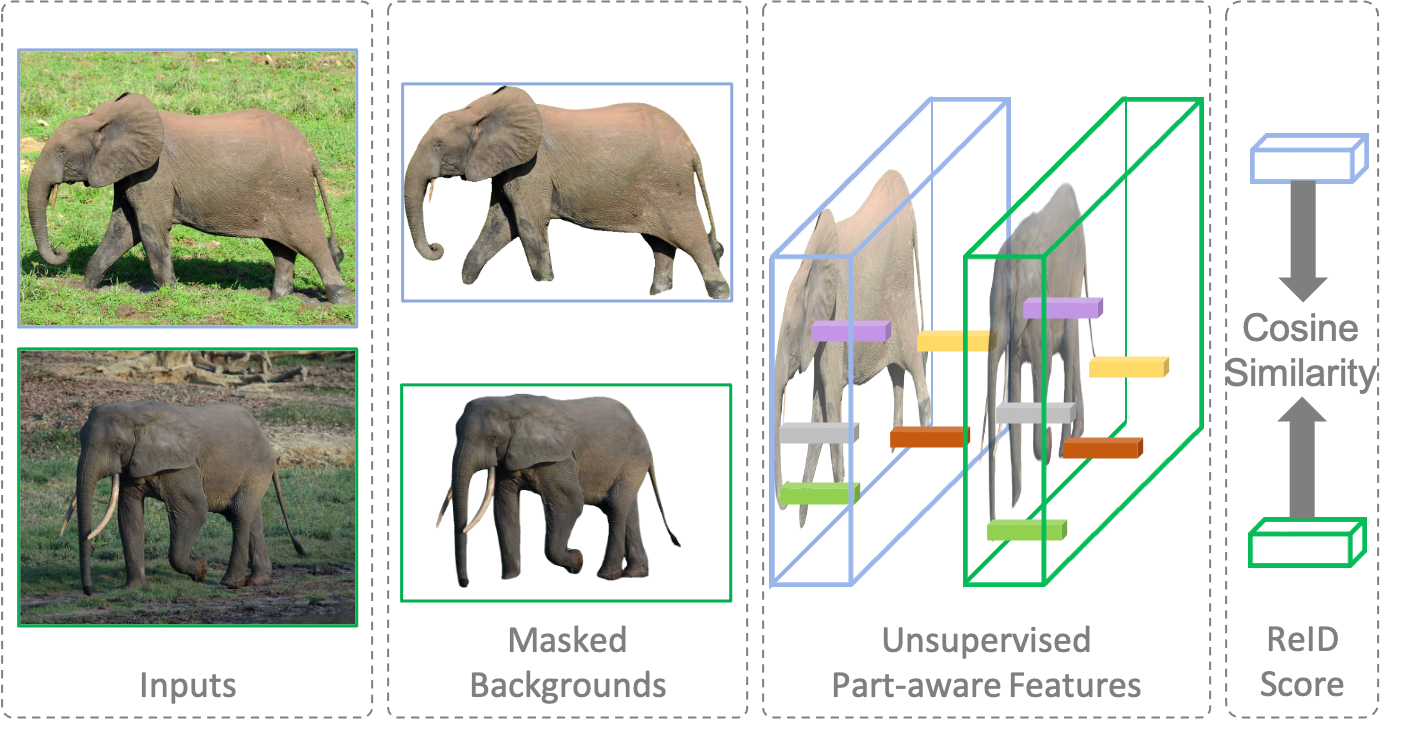}
 \caption{\textbf{Proposed Animal Re-ID Approach:} Addressing background bias in Re-ID models, our method masks out backgrounds to focus on the animal. It learns part-aware representations, ensuring consistency across subjects. Part-aware features are merged and a final Re-ID score is computed via cosine similarity.}
    \label{fig:intro}
  \end{figure}

Animal Re-ID faces unique challenges, including small non-diverse datasets leading to background bias, greater variations within an individual and smaller variations between different individuals compared to humans. These challenges necessitate animal-specific Re-ID methods.

This study focuses on Re-ID for elephants, yaks, and tigers, emphasizing the importance of feature alignment across pose variations due to their four-legged nature and the variability in their appearances. Unlike prior works \cite{liu_pose-guided_2019, liu_part-pose_2019, sun_beyond_2018,li2023automatic} that depend on pose annotations, our approach employs an unsupervised learning method~\cite{thewlis_unsupervised_2019} to identify semantically similar animal parts, enhancing Re-ID performance.

Our contributions illustrated in \cref{fig:intro} include: (i) Reducing background bias by creating and sharing background-free images, (ii) Utilizing an unsupervised method for learning part-based descriptors, thereby improving Re-ID accuracy without needing pose annotations and (iii) assessing the transferability of our method between different species, demonstrating the feasibility of transferring pose alignment across various species. 
\blfootnote{The code is made available in the following link: \url{https://github.com/Chloe-Yu/Animal-Re-ID}}

\section{Method}
\label{sec:model}

We present our approach for Animal Re-ID, starting with our technique for background removal to focus evaluation on animals rather than their surroundings. Next, we outline our Re-ID model's architecture and training approach. Lastly, we detail our integration of unsupervised landmark detection with our model to enable the learning of part-aware representations.

\begin{figure*}[t]
  \centering
  \includegraphics[width=0.8\linewidth]{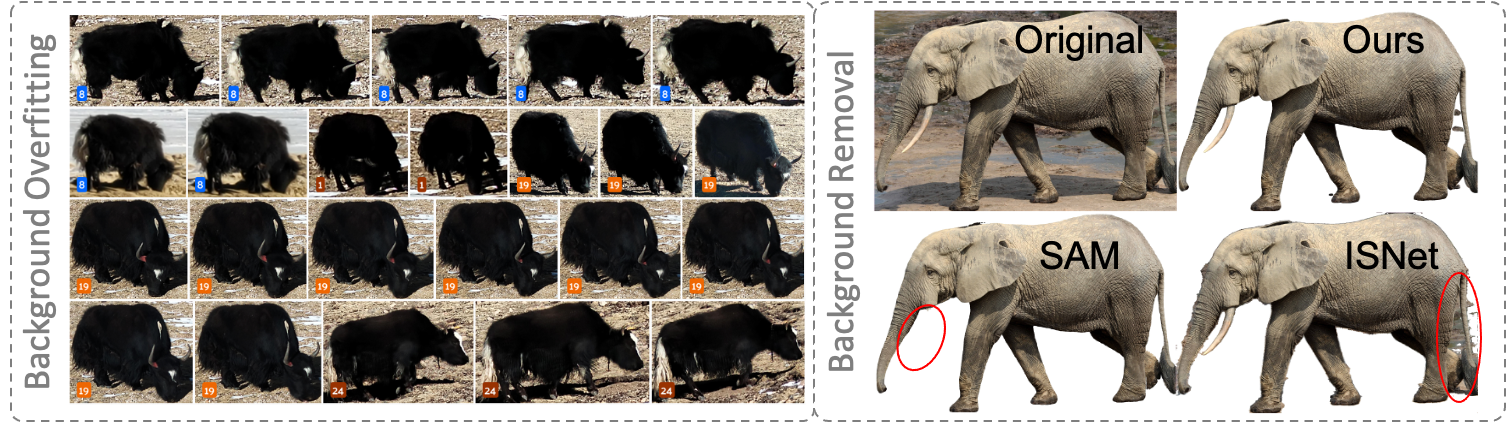}
\caption{\textbf{Bias Towards Background:} On the left, we display samples from the YakReID-103 dataset. Utilizing features from the PGCFL \cite{liu_pose-guided_2019} model, we identify the nearest neighbors. Each image's entity label is presented in the bottom-left corner. The retrieved images showcase four distinct entities, all sharing a remarkably similar background. This indicates that distances in the feature space are significantly influenced by background similarities. On the right, we exhibit the outcomes of our proposed background segmentation protocol. The top-left image is the original, the bottom-left depicts results from SAM, the bottom-right from ISNet, and the top-right combines outputs from both SAM and ISNet.}
  \label{fig:background_removal}
  \vspace{-0.5em}
\end{figure*}

\subsection{Background and Re-ID}
\label{sec:back_removal}
Background overfitting is a significant issue in animal Re-ID, exacerbated by small dataset sizes and repetitive backgrounds. It is particularly problematic in real-world scenarios where the same entity can be seen with lots of different backgrounds. We illustrate this problem on the left side of \cref{fig:background_removal}. 

To counter this, we diverge from previous strategies that adapt models to ignore backgrounds. Instead, we propose altering the data by removing backgrounds from all images using segmentation models ISNet \cite{isnet} and SAM \cite{kirillov2023segany}. We combine both models to produce refined segmentation masks, balancing ISNet's completeness with SAM's precision, particularly for distinguishing animal parts like tusks and horns. To get the best of both worlds, we combine both approaches to obtain foreground mask $M_{\ve{x}}$ from a given image $\ve{x}$ as :

\begin{equation}
M_{\ve{x}} = \bigcup_{}\{\ve{m}|\ve{m} \in S_{SAM}(\ve{x}) \land h(\ve{m};S_{ISNet}(\ve{x}))< \Theta)\}\;,
\end{equation}

\noindent where $S_{SAM}(x)$ are processed object masks from SAM, and $S_{ISNet}(x)$ is the output mask from ISNet. $h$ is a criterion (such as IoU) that filters $S_{SAM}(x)$ given $S_{ISNet}(x)$ with threshold $\Theta$. We provide additional details about the criterion in the supplementary material. 

The combination of both models yields better segmentation masks as shown in \cref{fig:background_removal}. In the following sections, we use this segmentation protocol to extract foreground masks for the different benchmarks and mask out the background both during training and evaluation.  Code and processed images will be made available to facilitate replication and further research.

\begin{figure*}[t]
  \centering
  \includegraphics[width=0.9\textwidth]{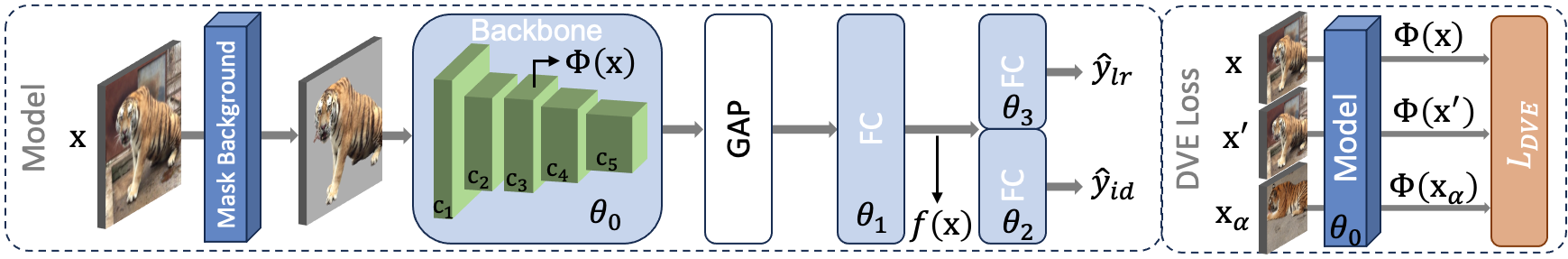}
\caption{\textbf{Overall architecture:} Left - proposed model's architecture. Input is a single image, processed through the first $3$ layers of backbone and a convolutional block to extract DVE features $\Phi(\ve x)$. Features produced after the fifth backbone layer continue to global average pooling and a linear layer for Re-ID features $f(\ve x)$, then pass through two classification heads for ID class and orientation prediction via linear layers and softmax operations. Right - $L_{DVE}$ for part-aware representation.}
  \label{fig:model}
    \vspace{-0.5em}
\end{figure*}


\subsection{Re-ID Approach}
\label{reid_method}

\subsubsection{Formalism}
In a standard Re-ID scenario, given a query image $\ve q$ of dimensions $w\times h\times 3$, the goal is to rank a gallery set of $N$ images $\mathcal{G}=\{\ve g_j\}_1^N$ to find those matching the query's identity. Images are ranked based on the cosine similarity between the query representation $f(\ve q) \in \R^{d}$ and each gallery image representation $f(\ve g_j) \in \R^{d}$, calculated as $c(f(\ve q) , f(\ve g_j))  = \frac{\langle f(\ve q) , f(\ve g_j)\rangle}{\|f(\ve q) \|\,\|f(\ve g_j)\|}$. Here, $f$ is the embedding function into a $d$-dimensional Re-ID space, parametrized by $\theta$. The training process focuses on optimizing $\theta$. The gallery encompasses one or multiple images of each unique animal identity, with a total of $C_{id}$ identities ($C_{id} <= N$).

\subsubsection{Architecture}

Consistent with prior Animal Re-ID studies \cite{liu_pose-guided_2019, zhang_yakreid-103_2021}, our model employs a modified SE-ResNet50 \cite{seresnet}, following adaptations in \cite{He2018BagOT}. We utilize the initial five convolutional blocks of SE-ResNet50. The output from the fifth block is processed through a Global Average Pooling (GAP) layer \cite{Lin2013NetworkIN}, condensing spatial dimensions into 1D vectors. These vectors are then passed through a sequence comprising a linear layer, batch normalization \cite{pmlr-v37-ioffe15}, and dropout \cite{srivastava_dropout_2014}, producing the Re-ID vector representation. For identity and orientation classification, the Re-ID features are further passed through two linear layers to produce the final logits.

Formally, for an input image $\ve x \in (0,255)^{W\times H\times 3}$, our Re-ID model operates as:

\begin{equation}
\label{eq:pipe}
\ve x \xmapsto{\text{BR}}
\ve{\bar x} \xmapsto{\text{SE}}
\ve F \xmapsto{\text{GAP}}
\ve G \xmapsto{\text{FC}_{\theta_1}}
f(\ve x) \xmapsto{\text{FC}_{\theta_{2:3}}}
\begin{aligned}
\hat{y}_{lr} \\
\hat{y}_{id}
\end{aligned} \;\; ,
\end{equation}

\noindent where $\text{BR}$ denotes the background removal step, producing the background-free image $\ve{\bar x}$. \text{SE} corresponds to the first 5 layers of a pre-trained SE-ResNet50. $F$ in $\R^{w\times h\times d_{se}}$ correspond to the activation of the fifth layer. $G$ in $\R^{d_g}$ is the output of the Global Average Pooling operation. 
$f(\ve x) \in \R^{d}$ is the Re-ID representation, and $\hat{y}_{lr} \in \R^{1}$ and $\hat{y}_{id} \in \R^{C_{id}}$ are orientation and identity classification logits respectively. The complete architecture is shown in \cref{fig:model}.


\subsection{Part-aware Feature Learning}
Identifying individuals within the same species depends on fine body part details, such as tigers' stripes \cite{tiger_stripe}, yaks' horns and fur, or elephants' ears and tusks. Part-based Re-ID methods, which are effective \cite{liu_part-pose_2019,li_atrw_2020}, usually need pose annotations. Our unsupervised method leverages the Descriptor Vector Exchange (DVE) for learning part-specific features without posture labels, using DVE's technique for unsupervised dense landmark prediction \cite{thewlis_unsupervised_2019}. 

DVE's local image descriptors are designed to be equivariant to transformations and invariant to variations within a category. The DVE objective function is:

\begin{equation}
\begin{aligned}
L_{DVE} = \frac{1}{|\Omega|^2} \int_{\Omega} \int_{\Omega}\|v-g u\| p(v|u; \Phi, \ve x, \ve{x'},\ve{x_{\alpha}}) d u d v    
\end{aligned}\;,
\label{eq:dve_loss}
\end{equation}

\noindent where $\Phi$ is a projection from the image domain to the local descriptors domain, $g$ is a random warping function, $\ve x$ is an image, and $\ve{x'}$ = $g \ve x$ its deformation. $p(v|u; \Phi, \ve x, \ve{x'},\ve{x_{\alpha}})$ is the probability of pixel $u$ in image $\ve x$ matching with
pixel $v$ in image $\ve{x'}$. The probability computation uses an auxiliary image $\ve{x_{\alpha}}$  to make the local descriptor invariant to intra-category variations. For more information about the DVE objective function please refer to \cite{thewlis_unsupervised_2019}.

\paragraph{DVE for Re-ID}
DVE descriptors are invariant to intra-category variations, meaning similar parts across different subjects (e.g. left leg of a tiger) will have comparable descriptors. Leveraging this, we incorporate the DVE property into our Re-ID model by using the DVE loss $L_{DVE}$ to refine our model's features. Specifically, the activation of the third convolutional layer of the SE-ResNet is fed into an extra convolution layer to get the descriptors $\Phi(\ve x)$ on which $L_{DVE}$ is applied.

Formally, those local descriptors are the results of the following steps:
\begin{equation}
\ve x \xmapsto{\text{BR}}
\ve{\bar x} \xmapsto{\text{SE}_{c_{1:3}}}
\ve{ F'} \xmapsto{\text{CONV}}
\Phi(\ve{\bar x})\;,
\label{eq:pipe2}
\end{equation}

\noindent $\ve{ F'}$ in $\R^{w'\times h'\times d_{se}'}$ correspond to the activation of the third layer. $\Phi(\ve{\bar x})$ in $\R^{w'\times h'\times d_{dve}}$ is the DVE feature.

The output resolution of our model's backbone is reduced by a factor of 8 compared to the input image. Given that the resolution post-5th layer is too low for effective local descriptor learning via DVE, we opt to apply the DVE objective at a higher resolution stage, specifically after the third layer of the SE-ResNet, where the output is only downscaled by a factor of 4. \cref{fig:model} illustrates how $L_{DVE} $ is obtained. 

\subsection{Loss Function and Training}

In addition to $L_{DVE}$, our model incorporates standard Re-ID losses similar to those in prior studies \cite{liu_pose-guided_2019, zhang_yakreid-103_2021}. The loss function comprises two classification losses—one for orientation and another for identity—along with a loss for learning the Re-ID representation. The classification losses read as follows:

\begin{equation}
L_{ID} = -  \sum_{c=1}^{C_{id}} y_{id}^{c} \log(\hat{y}_{id}^{c})\;,
\label{eq:id_loss}
\end{equation}

\begin{equation}
L_{LR} =  - (y_{lr} \log(\hat{y}_{lr}) + (1 - y_{lr}) \log(1 - \hat{y}_{lr}))\;,
\label{eq:lr_loss}
\end{equation}

\noindent where $C_{id}$ is the number of entity classes, $y_{id}$ and $y_{lr}$ is the ground truth label for entity and orientation, respectively. 

For learning the Re-ID representation, we diverge from the common use of triplet loss in previous studies and instead utilize circle loss \cite{circle_loss}. Given a sample $\ve{x}$, let's consider $K$ within-class and $L$ between-class similarity scores, denoted by ${s_{p}^{i}}(i = 1, 2, ... , K)$ and ${s_{j}^{n}}(j = 1, 2, ... , L)$, respectively. The formulation of the circle loss is as follows:

\begin{equation}
\begin{split}
L_{reID} = & \log\left[ \sum_{j=1}^{L} \exp\left(\gamma [s_{n}^{j}+m]_{+}\times(s_{n}^{j}-m)\right) \right. \\
& \times \left. \sum_{i=1}^{K} \exp\left(-\gamma [-s_{p}^{i}+1+m]_{+}\times(s_{p}^{i}-1+m)\right) \right],
\end{split}
\label{eq:reid_loss}
\end{equation}

\noindent where $\gamma$ is the scale factor and $m$ is the margin.

The overall objective function used during training is as follows:

\begin{equation}
L = L_{ID} + L_{LR} +\lambda_{reID} \times  L_{reID}+ \lambda_{DVE} \times L_{DVE}\;,
\label{eq:total_loss}
\end{equation}

\noindent where $\lambda_{reID}$ and $\lambda_{DVE}$ are hyperparameters allowing to tune respectively the strength of $L_{reID}$ and of the regularization effect of $L_{DVE}$. 
 
\begin{table}[t]
  \centering
  \begin{tabular}{ccccc}
    \toprule
    \multicolumn{2}{c}{Masked Backgrounds} & \multicolumn{3}{c}{ATRW} \\
    \cmidrule(lr{.75em}){1-2} \cmidrule(lr{.75em}){3-5}
    Train & Test & mmAP & R@1(s) & R@1(c) \\
    \midrule   
     &  &74.5 & 95.7  & 90.3  \\    
     & \checkmark & 60.2 & 88.6  & 83.4  \\
    \checkmark & \checkmark &  66.9 & 90.8  & 86.3  \\
    \checkmark &  & 73.4 & 96.9  & 89.7  \\ 
    \bottomrule
  \end{tabular}

\caption{\textbf{Background Bias Measurement} Using the ATRW dataset, we assessed background bias in the PGCFL re-ID model. Training on original images led to performance collapse with masked backgrounds, highlighting benchmark bias and its effect on the model. Training on images with masked backgrounds improved model performance, which further enhanced with background inclusion.}
\label{tab:background_eval}
\end{table}

\section{Experiments}
\label{sec:experiments}

In the following, we first describe the datasets, metrics and training procedures. Then, we investigate the effect of background removal on Re-ID performance. Finally, we show the benefit of the proposed part-aware features through a series of quantitative and qualitative evaluations.

\paragraph{Datasets}\label{data}
The ATRW dataset \cite{li_atrw_2020} is the largest wildlife Re-ID dataset, featuring 182 entities across 92 tigers, with a training set of 1,887 images from 107 entities and a test set of 1,762 images from 75 entities, without utilizing provided pose annotations. YakReID-103 \cite{zhang_yakreid-103_2021} includes 1,404 training images of 121 entities, and we only use the hard-testing subset of 433 images, where similar poses and backgrounds are excluded. ELPephants \cite{korschens_elpephants_2019} contains 2,078 images of 276 elephants. We manually completed
the dataset’s partial heading direction annotations and included only entities with multiple side-view images. It's split into 1,380 training and 380 testing images without predefined bounds. Each dataset ensures no entity overlap between training and testing, considering each side of an individual as a distinct entity.

\paragraph{Metrics}
The evaluation employs mean average precision (mAP) and recall at 1 (R@1), with AP calculated per query from ranked gallery lists. For ATRW, metrics are separately computed for single-camera (R@1(s)) and cross-camera (R@1(c)) settings, with mmAP as their average. YakReID-103 and ELPephants lack camera data, considering any same-entity gallery image as positive.

\paragraph{Implementation details}
Utilizing SE-ResNet50 pre-trained on ImageNet as the backbone, similar to PGCFL and PPGNet, our model is developed in Pytorch and operates on an Nvidia A100 GPU. Images are resized to 224x224, with data augmentation including random cropping, patch erasing, and flipping, adjusting orientation labels accordingly. Training ceases after 80 epochs for ATRW and YakReID-103, and 100 for ELPephants, starting with a fixed backbone for the initial three epochs. The learning rate is set at $0.001$ for the backbone and $0.01$ for other layers, reduced by tenfold after two-thirds of the epochs. SGD optimizer is used with momentum $0.9$, weight decay $5e-4$, label smoothing $0.1$, and a batch size of $30$. Parameters $d_{DVE}$, $\lambda_{reID}$, and $\lambda_{DVE}$ are set to 64, 2, and $0.2$ respectively, employing Circle Loss with $\gamma$ of $64$ and $m$ of $0.25$.

For testing, features from the original and flipped images are concatenated for the Re-ID score, incorporating a re-ranking strategy for ATRW results. Despite not being discussed in the original works, we experimented with a vanilla triplet sampling strategy for batch processing, ensuring adequate negative samples for Circle Loss.

\begin{figure}[t]
  \centering
  \includegraphics[width=0.85\linewidth]{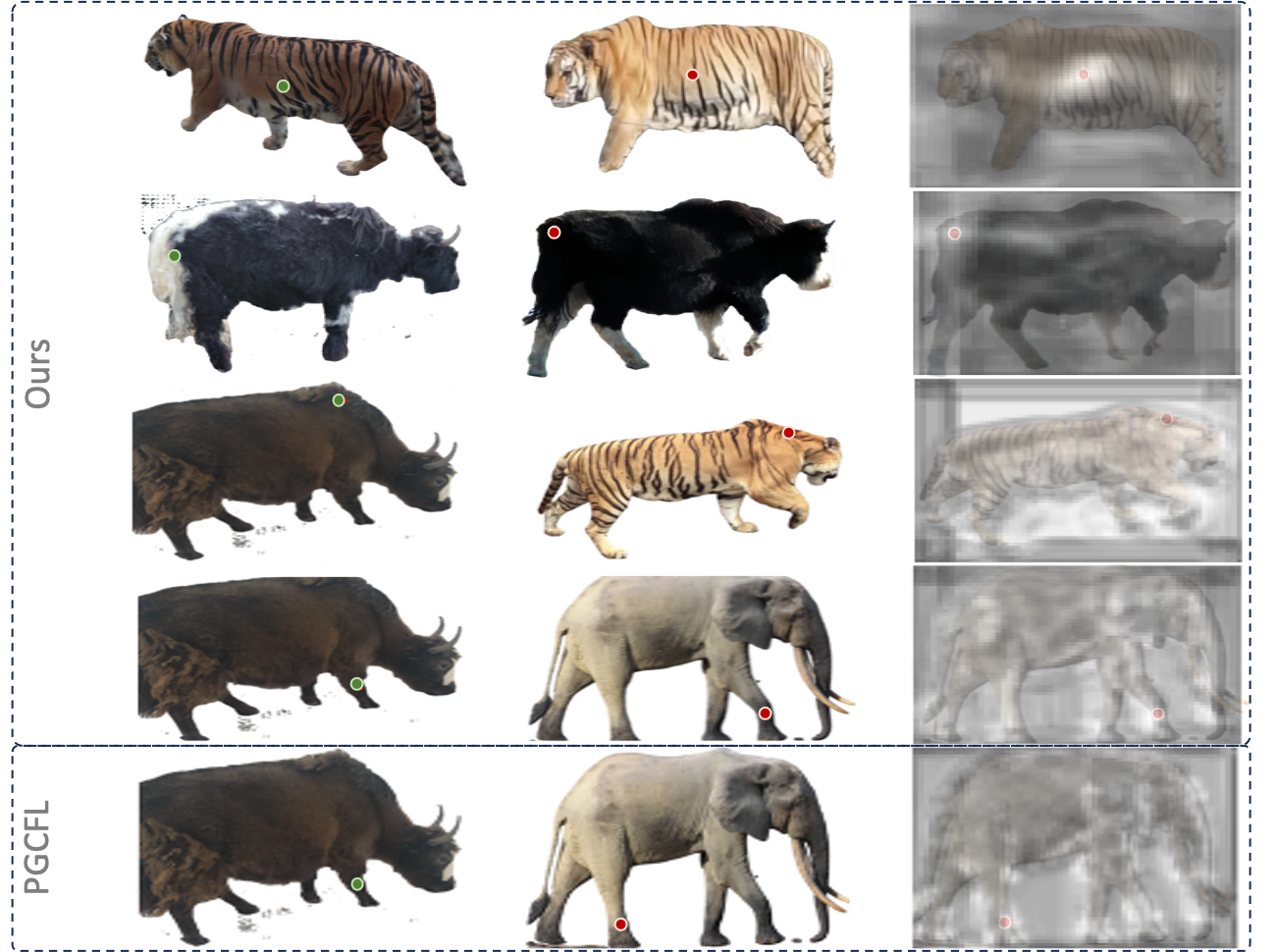}
  \caption{\textbf{Visualization of the feature learned with $L_{DVE}$} The first two rows show intra-species part alignment, the next two rows demonstrate that a model trained \emph{solely} on tiger can generalize to other species and maintain alignment even in inter-species scenario. The final row is the results from the PGCFL baseline. 
  In each row, the green dot in the left image is the local query, while the red dot in the center image indicates its matching point. The right-most image provides a heatmap overlaid on the target image, showcasing the similarities between the local query and the center image.}
  \label{fig:dve_visualization}
\end{figure}

\begin{table*}[ht]
  \centering
  \begin{tabular}{lccccccccc}
    \toprule
    & & & \multicolumn{3}{c}{ATRW} & \multicolumn{2}{c}{YakReID-103} & \multicolumn{2}{c}{ELPephants} \\
    \cmidrule(lr{.75em}){4-6}\cmidrule(lr{.75em}){7-8}\cmidrule(lr{.75em}){9-10}
    Methods & Org. Re-ID Task & Pose GT  & mmAP & R@1(s) & R@1(c) & mAP & R@1 & mAP & R@1 \\
    \midrule
    \cmidrule(lr{.75em}){4-10}
    CLIP-Re-ID ViT \cite{li2023clip} & Person  & \xmark & 56.5 & 86.3 & 72.0 & 49.8 & 82.2 & 12.7 & 25.3 \\
    PPGNet R-50 \cite{liu_part-pose_2019}  & Animal  & \checkmark & 68.3 & 81.2 & 81.1 & - &- & - &-\\
    ResNet50 \cite{DBLP:journals/corr/HeZRS15}  & Animal  & \xmark & 65.9 & 91.1 & 83.4& 60.9 & 86.0 & 20.0 & 33.9 \\
    ViT \cite{dosovitskiy2020image} & Animal  & \xmark & 65.5 & 90.3  & 79.4 &\textbf{61.3}& 88.4 & 20.6 & 36.8  \\
    PGCFL \cite{liu_pose-guided_2019} & Animal   & \xmark & 66.9 & 90.8  & \textbf{86.3} & 55.8 & 82.7 & 18.5 & 33.4  \\
    Ours & Animal   & \xmark & \textbf{68.6} & \textbf{92.0} & 84.6 & 61.0 & \textbf{89.4} & \textbf{24.3} & \textbf{38.7}  \\
    \midrule
  \end{tabular}

\caption{\textbf{Comparison to State-of-the-Art:} Our method is evaluated on three datasets, each representing a different species: ATRW (Tiger), YakReID-103 (Yak), and ELPephants (Elephants), originally proposed for various Re-ID tasks. Results from images with masked backgrounds, detailed in \cref{sec:back_removal}, are highlighted. Our model achieves top performance, surpassing existing baselines in mAP on ATRW and ELPephants, even outperforming PPGNet, which utilizes extra pose labels. For results on original images see supplementary material.}
  \vspace{-0.5em}
\label{tab:sota}
\end{table*}

\paragraph{Effect of Background Removal}
As can be seen in \cref{tab:background_eval} background removal significantly impacts performance, with a mmAP decrease of $10.2\%$ when training and testing without backgrounds, versus a $19.2\%$ drop when testing a full-image trained model on a masked background. However, models trained on masked backgrounds can still leverage backgrounds in testing, with an observed $9.7\%$ mmAP increase, suggesting the importance of background-independent Re-ID models. Subsequent results are reported using masked backgrounds to emphasize identity recognition regardless of the background.

\paragraph{Comparison to State-of-the-Art}

Our method surpasses existing models like PGCFL \cite{liu_pose-guided_2019}, PPGNet \cite{liu_part-pose_2019} (ATRW challenge winner), CLIP-Re-ID \cite{li2023clip} (a recent person Re-ID method), and baselines with ResNet50 and ViT backbones using $L_{base} = L_{ID} + L_{LR} + \lambda_{reID} \times L_{reID}$. Specifically, PPGNet, which depends on ground-truth poses, is evaluated solely on ATRW. Results can be found in \cref{tab:sota}. Our approach leads in mmAP and R@1(s) on ATRW, equals ViT in mAP, and excels in R@1 on YakReID-103, while significantly outperforming all on ELPephants by a 5.8 mAP margin against PGCFL. CLIP-Re-ID's adaptation to animal datasets is hindered by greater intra-identity variations in animals, a challenge not as prevalent in human subjects. Although PCN-RERP shows promising mAP on YakReID-103, it's limited to datasets featuring animals in standing postures. Results without background masking are provided in supplementary material.

\paragraph{Qualitative Analysis of DVE.}
DVE helps the model embed body part information in its activation. Features for a given body part will share similarity across different entities. This helps align body parts and thus facilitates Re-ID. As shown in the two top rows of \cref{fig:dve_visualization}, when visualizing the descriptors $\Phi(\ve x)$ learn through the loss $L_{DVE}$ we can successfully match body parts.

\paragraph{Towards transfer between species.}
We also investigate the possibility of transfer between species. Using a model trained on tigers, we visualize third-layer features of the Re-ID backbone on yaks and elephants. While tigers, yaks, and elephants have widely different shapes and being only trained on tigers, \comment{\emph{only}}our model was able to match body parts across species as seen in rows 3 and 4 of \cref{fig:dve_visualization}.

\subsection{Limitations}

\paragraph{Background removal.} 
There are cases where both ISNet and SAM fail. 
In these cases, the combination of the two methods does not help. But these are rare cases and are usually linked to low image quality.

\paragraph{Inter-species transfer.} 
While the proposed approach showed promising results on inter-species transfer, it also struggled in some specific cases. A model trained only on tigers has difficulty matching horns between yaks or tusks between elephants. We further illustrate these failure cases in the supplementary material.

\section{Conclusion}
\label{sec:conclusion}
We have introduced a novel method to advance animal re-identification. Our approach diverges from prior work by learning part-aware features in an unsupervised manner. Furthermore, by automatically masking the background, we not only address a recurrent challenge in animal Re-ID benchmarks but also effectively reduce the model's inclination to overly focus on the background. Lastly, we demonstrated that our approach represents an initial step towards the inter-species transferability of Re-ID models. In future studies, we plan to further refine our method's inter-species generalization capabilities by leveraging additional multi-species unsupervised training.

{
    \small
    \bibliographystyle{ieeenat_fullname}
    \bibliography{main}

\begin{thebibliography}{33}
\providecommand{\natexlab}[1]{#1}
\providecommand{\url}[1]{\texttt{#1}}
\expandafter\ifx\csname urlstyle\endcsname\relax
  \providecommand{\doi}[1]{doi: #1}\else
  \providecommand{\doi}{doi: \begingroup \urlstyle{rm}\Url}\fi

\bibitem[Chen et~al.(2020)Chen, Liu, Wu, and Chien]{chen2020orientation}
Tsai-Shien Chen, Chih-Ting Liu, Chih-Wei Wu, and Shao-Yi Chien.
\newblock Orientation-aware vehicle re-identification with semantics-guided part attention network.
\newblock In \emph{Computer Vision--ECCV 2020: 16th European Conference, Glasgow, UK, August 23--28, 2020, Proceedings, Part II 16}, pages 330--346. Springer, 2020.

\bibitem[Cheng et~al.(2022)Cheng, Wang, and Han]{isnet}
Gong Cheng, Guangxing Wang, and Junwei Han.
\newblock Isnet: Towards improving separability for remote sensing image change detection.
\newblock \emph{IEEE Transactions on Geoscience and Remote Sensing}, 60:\penalty0 1--11, 2022.

\bibitem[Dosovitskiy et~al.(2020)Dosovitskiy, Beyer, Kolesnikov, Weissenborn, Zhai, Unterthiner, Dehghani, Minderer, Heigold, Gelly, et~al.]{dosovitskiy2020image}
Alexey Dosovitskiy, Lucas Beyer, Alexander Kolesnikov, Dirk Weissenborn, Xiaohua Zhai, Thomas Unterthiner, Mostafa Dehghani, Matthias Minderer, Georg Heigold, Sylvain Gelly, et~al.
\newblock An image is worth 16x16 words: Transformers for image recognition at scale.
\newblock \emph{arXiv preprint arXiv:2010.11929}, 2020.

\bibitem[He et~al.(2022)He, Chen, Liu, Kortylewski, Yang, Bai, and Wang]{he2022transfg}
Ju He, Jie-Neng Chen, Shuai Liu, Adam Kortylewski, Cheng Yang, Yutong Bai, and Changhu Wang.
\newblock Transfg: A transformer architecture for fine-grained recognition.
\newblock In \emph{Proceedings of the AAAI Conference on Artificial Intelligence}, pages 852--860, 2022.

\bibitem[He et~al.(2016)He, Zhang, Ren, and Sun]{DBLP:journals/corr/HeZRS15}
Kaiming He, Xiangyu Zhang, Shaoqing Ren, and Jian Sun.
\newblock Deep residual learning for image recognition.
\newblock In \emph{Proceedings of the IEEE conference on computer vision and pattern recognition}, pages 770--778, 2016.

\bibitem[He et~al.(2018)He, Liang, Li, and Sun]{he2018deep}
Lingxiao He, Jian Liang, Haiqing Li, and Zhenan Sun.
\newblock Deep spatial feature reconstruction for partial person re-identification: Alignment-free approach.
\newblock In \emph{Proceedings of the IEEE conference on computer vision and pattern recognition}, pages 7073--7082, 2018.

\bibitem[He et~al.(2021)He, Luo, Wang, Wang, Li, and Jiang]{he_transreid_2021}
Shuting He, Hao Luo, Pichao Wang, Fan Wang, Hao Li, and Wei Jiang.
\newblock Transreid: Transformer-based object re-identification.
\newblock In \emph{Proceedings of the IEEE/CVF international conference on computer vision}, pages 15013--15022, 2021.

\bibitem[Hiby et~al.(2009)Hiby, Lovell, Patil, Kumar, Gopalaswamy, and Karanth]{tiger_stripe}
Lex Hiby, Phil Lovell, Narendra Patil, N~Samba Kumar, Arjun~M Gopalaswamy, and K~Ullas Karanth.
\newblock A tiger cannot change its stripes: using a three-dimensional model to match images of living tigers and tiger skins.
\newblock \emph{Biology letters}, 5\penalty0 (3):\penalty0 383--386, 2009.

\bibitem[Hu et~al.(2018)Hu, Shen, and Sun]{seresnet}
Jie Hu, Li Shen, and Gang Sun.
\newblock Squeeze-and-excitation networks.
\newblock In \emph{Proceedings of the IEEE conference on computer vision and pattern recognition}, pages 7132--7141, 2018.

\bibitem[Ioffe and Szegedy(2015)]{pmlr-v37-ioffe15}
Sergey Ioffe and Christian Szegedy.
\newblock Batch normalization: Accelerating deep network training by reducing internal covariate shift.
\newblock In \emph{International conference on machine learning}, pages 448--456. pmlr, 2015.

\bibitem[Khorramshahi et~al.(2020)Khorramshahi, Peri, Chen, and Chellappa]{khorramshahi2020devil}
Pirazh Khorramshahi, Neehar Peri, Jun-cheng Chen, and Rama Chellappa.
\newblock The devil is in the details: Self-supervised attention for vehicle re-identification.
\newblock In \emph{Computer Vision--ECCV 2020: 16th European Conference, Glasgow, UK, August 23--28, 2020, Proceedings, Part XIV 16}, pages 369--386. Springer, 2020.

\bibitem[Kirillov et~al.(2023)Kirillov, Mintun, Ravi, Mao, Rolland, Gustafson, Xiao, Whitehead, Berg, Lo, et~al.]{kirillov2023segany}
Alexander Kirillov, Eric Mintun, Nikhila Ravi, Hanzi Mao, Chloe Rolland, Laura Gustafson, Tete Xiao, Spencer Whitehead, Alexander~C Berg, Wan-Yen Lo, et~al.
\newblock Segment anything.
\newblock \emph{arXiv preprint arXiv:2304.02643}, 2023.

\bibitem[Korschens and Denzler(2019)]{korschens_elpephants_2019}
Matthias Korschens and Joachim Denzler.
\newblock Elpephants: A fine-grained dataset for elephant re-identification.
\newblock In \emph{Proceedings of the IEEE/CVF International Conference on Computer Vision Workshops}, pages 0--0, 2019.

\bibitem[Li et~al.(2023{\natexlab{a}})Li, Zhang, Cuo, Zhao, Zhou, and Jiancuo]{li2023automatic}
Lei Li, Tingting Zhang, Da Cuo, Qijun Zhao, Liyuan Zhou, and Suonan Jiancuo.
\newblock Automatic identification of individual yaks in in-the-wild images using part-based convolutional networks with self-supervised learning.
\newblock \emph{Expert Systems with Applications}, 216:\penalty0 119431, 2023{\natexlab{a}}.

\bibitem[Li et~al.(2019)Li, Li, Tang, Qian, and Lin]{li_atrw_2020}
Shuyuan Li, Jianguo Li, Hanlin Tang, Rui Qian, and Weiyao Lin.
\newblock Atrw: a benchmark for amur tiger re-identification in the wild.
\newblock \emph{arXiv preprint arXiv:1906.05586}, 2019.

\bibitem[Li et~al.(2023{\natexlab{b}})Li, Sun, and Li]{li2023clip}
Siyuan Li, Li Sun, and Qingli Li.
\newblock Clip-reid: exploiting vision-language model for image re-identification without concrete text labels.
\newblock In \emph{Proceedings of the AAAI Conference on Artificial Intelligence}, pages 1405--1413, 2023{\natexlab{b}}.

\bibitem[Lin et~al.(2013)Lin, Chen, and Yan]{Lin2013NetworkIN}
Min Lin, Qiang Chen, and Shuicheng Yan.
\newblock Network in network.
\newblock \emph{CoRR}, abs/1312.4400, 2013.

\bibitem[Liu et~al.(2019{\natexlab{a}})Liu, Zhang, and Guo]{liu_part-pose_2019}
Cen Liu, Rong Zhang, and Lijun Guo.
\newblock Part-pose guided amur tiger re-identification.
\newblock In \emph{Proceedings of the IEEE/CVF International Conference on Computer Vision Workshops}, pages 0--0, 2019{\natexlab{a}}.

\bibitem[Liu et~al.(2019{\natexlab{b}})Liu, Zhao, Zhang, Cheng, and Zhu]{liu_pose-guided_2019}
Ning Liu, Qijun Zhao, Nan Zhang, Xinhua Cheng, and Jianing Zhu.
\newblock Pose-guided complementary features learning for amur tiger re-identification.
\newblock In \emph{Proceedings of the IEEE/CVF international conference on computer vision workshops}, pages 0--0, 2019{\natexlab{b}}.

\bibitem[Liu et~al.(2020)Liu, Huang, and Zhang]{LIU2020102018}
Shuang Liu, Wenmin Huang, and Zhong Zhang.
\newblock Person re-identification using hybrid task convolutional neural network in camera sensor networks.
\newblock \emph{Ad Hoc Networks}, 97:\penalty0 102018, 2020.

\bibitem[Luo et~al.(2019)Luo, Gu, Liao, Lai, and Jiang]{He2018BagOT}
Hao Luo, Youzhi Gu, Xingyu Liao, Shenqi Lai, and Wei Jiang.
\newblock Bag of tricks and a strong baseline for deep person re-identification.
\newblock \emph{2019 IEEE/CVF Conference on Computer Vision and Pattern Recognition Workshops (CVPRW)}, pages 1487--1495, 2019.

\bibitem[Martin et~al.(2007)Martin, Kitchens, and Hines]{Martin2007ImportanceOW}
Julien Martin, Wiley~M. Kitchens, and James~E Hines.
\newblock Importance of well‐designed monitoring programs for the conservation of endangered species: Case study of the snail kite.
\newblock \emph{Conservation Biology}, 21, 2007.

\bibitem[Radford et~al.(2021)Radford, Kim, Hallacy, Ramesh, Goh, Agarwal, Sastry, Askell, Mishkin, Clark, et~al.]{radford2021learning}
Alec Radford, Jong~Wook Kim, Chris Hallacy, Aditya Ramesh, Gabriel Goh, Sandhini Agarwal, Girish Sastry, Amanda Askell, Pamela Mishkin, Jack Clark, et~al.
\newblock Learning transferable visual models from natural language supervision.
\newblock In \emph{International conference on machine learning}, pages 8748--8763. PMLR, 2021.

\bibitem[Srivastava et~al.(2014)Srivastava, Hinton, Krizhevsky, Sutskever, and Salakhutdinov]{srivastava_dropout_2014}
Nitish Srivastava, Geoffrey Hinton, Alex Krizhevsky, Ilya Sutskever, and Ruslan Salakhutdinov.
\newblock Dropout: a simple way to prevent neural networks from overfitting.
\newblock \emph{The journal of machine learning research}, 15\penalty0 (1):\penalty0 1929--1958, 2014.

\bibitem[Sun et~al.(2018)Sun, Zheng, Yang, Tian, and Wang]{sun_beyond_2018}
Yifan Sun, Liang Zheng, Yi Yang, Qi Tian, and Shengjin Wang.
\newblock Beyond part models: Person retrieval with refined part pooling (and a strong convolutional baseline).
\newblock In \emph{Proceedings of the European conference on computer vision (ECCV)}, pages 480--496, 2018.

\bibitem[Sun et~al.(2020)Sun, Cheng, Zhang, Zhang, Zheng, Wang, and Wei]{circle_loss}
Yifan Sun, Changmao Cheng, Yuhan Zhang, Chi Zhang, Liang Zheng, Zhongdao Wang, and Yichen Wei.
\newblock Circle loss: A unified perspective of pair similarity optimization.
\newblock In \emph{Proceedings of the IEEE/CVF conference on computer vision and pattern recognition}, pages 6398--6407, 2020.

\bibitem[Thewlis et~al.(2019)Thewlis, Albanie, Bilen, and Vedaldi]{thewlis_unsupervised_2019}
James Thewlis, Samuel Albanie, Hakan Bilen, and Andrea Vedaldi.
\newblock Unsupervised learning of landmarks by descriptor vector exchange.
\newblock In \emph{Proceedings of the IEEE/CVF International Conference on Computer Vision}, pages 6361--6371, 2019.

\bibitem[Wang et~al.(2018)Wang, Yuan, Chen, Li, and Zhou]{wang2018learning}
Guanshuo Wang, Yufeng Yuan, Xiong Chen, Jiwei Li, and Xi Zhou.
\newblock Learning discriminative features with multiple granularities for person re-identification.
\newblock In \emph{Proceedings of the 26th ACM international conference on Multimedia}, pages 274--282, 2018.

\bibitem[Wang et~al.(2020)Wang, Lai, Liang, and Wang]{wang2020smoothing}
Guangcong Wang, Jian-Huang Lai, Wenqi Liang, and Guangrun Wang.
\newblock Smoothing adversarial domain attack and p-memory reconsolidation for cross-domain person re-identification.
\newblock In \emph{Proceedings of the IEEE/CVF conference on computer vision and pattern recognition}, pages 10568--10577, 2020.

\bibitem[Wu et~al.(2019)Wu, Zheng, Zhang, Yuan, Cheng, Zhao, Lin, Zhao, Jiang, and Huang]{wu_deep_2019}
Di Wu, Si-Jia Zheng, Xiao-Ping Zhang, Chang-An Yuan, Fei Cheng, Yang Zhao, Yong-Jun Lin, Zhong-Qiu Zhao, Yong-Li Jiang, and De-Shuang Huang.
\newblock Deep learning-based methods for person re-identification: A comprehensive review.
\newblock \emph{Neurocomputing}, 337:\penalty0 354--371, 2019.

\bibitem[Ye et~al.(2021)Ye, Shen, Lin, Xiang, Shao, and Hoi]{ye_deep_2022}
Mang Ye, Jianbing Shen, Gaojie Lin, Tao Xiang, Ling Shao, and Steven~CH Hoi.
\newblock Deep learning for person re-identification: A survey and outlook.
\newblock \emph{IEEE transactions on pattern analysis and machine intelligence}, 44\penalty0 (6):\penalty0 2872--2893, 2021.

\bibitem[Zhang et~al.(2021)Zhang, Zhao, Da, Zhou, Li, and Jiancuo]{zhang_yakreid-103_2021}
Tingting Zhang, Qijun Zhao, Cuo Da, Liyuan Zhou, Lei Li, and Suonan Jiancuo.
\newblock Yakreid-103: A benchmark for yak re-identification.
\newblock In \emph{2021 IEEE International Joint Conference on Biometrics (IJCB)}, pages 1--8. IEEE, 2021.

\bibitem[Zhang et~al.(2018)Zhang, Zhou, Lin, and Sun]{zhang2018shufflenet}
Xiangyu Zhang, Xinyu Zhou, Mengxiao Lin, and Jian Sun.
\newblock Shufflenet: An extremely efficient convolutional neural network for mobile devices.
\newblock In \emph{Proceedings of the IEEE conference on computer vision and pattern recognition}, pages 6848--6856, 2018.

\end{thebibliography}
}

\renewcommand\thefigure{A.\arabic{figure}} 
\renewcommand\thetable{A.\arabic{table}} 
\renewcommand\thesection{A.\arabic{section}}

 \clearpage
\setcounter{page}{1}
\maketitlesupplementary

In the supplementary material accompanying this paper, we offer enhanced visualizations that highlight the complexities involved in animal re-identification. Additionally, we conduct a comprehensive ablation study to assess the impact of each component within our proposed model. We also present qualitative results to demonstrate the model's performance, provide an in-depth analysis of the model's capability to transfer knowledge across different species, and furnish supplementary details regarding the implementation. Our aim is to provide a thorough understanding of our approach and its underlying mechanisms.

\section{Challenge of Animal Re-ID}
\begin{figure}[h]
  \centering
  \includegraphics[width=\linewidth]{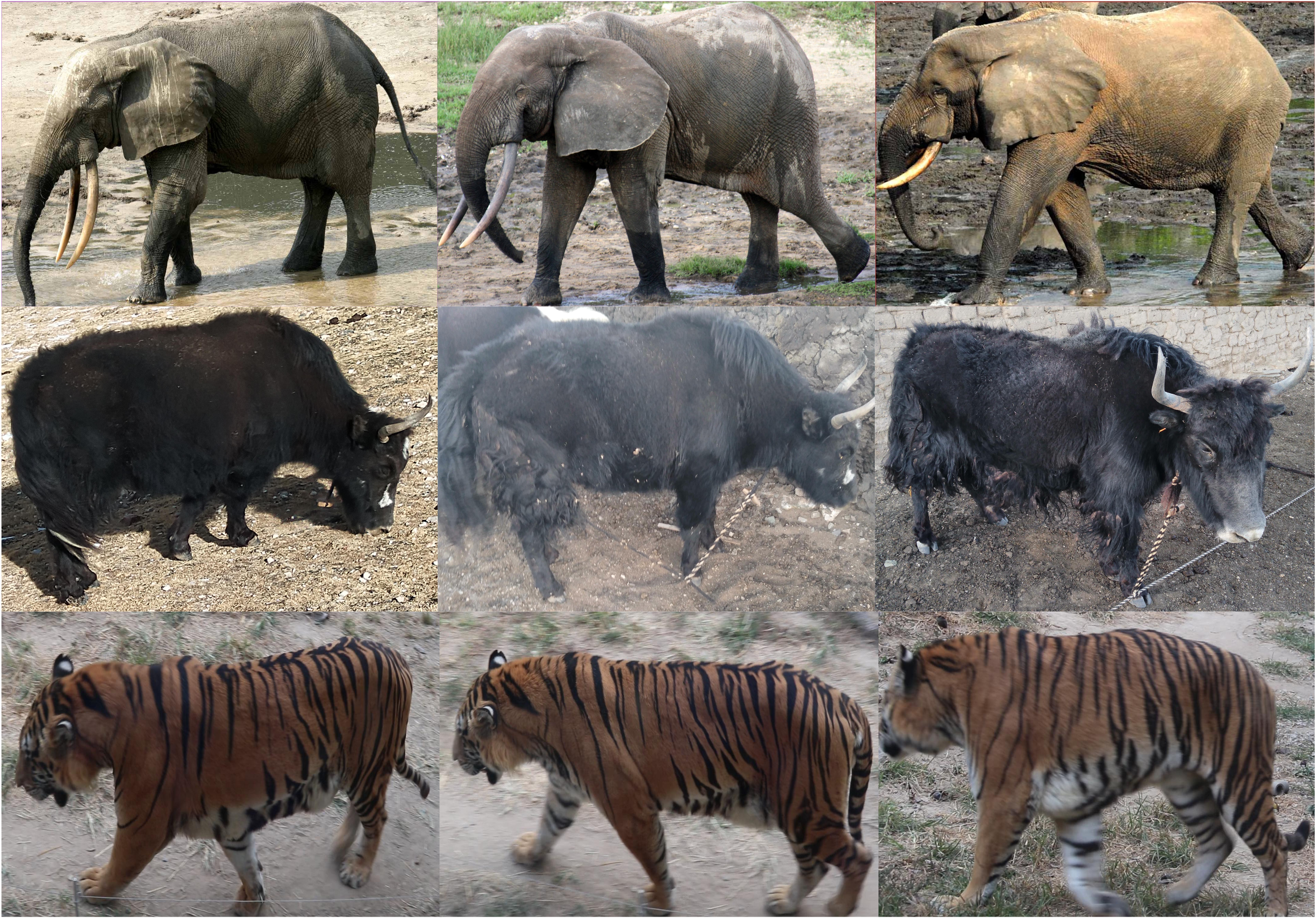}
  \caption{\textbf{Sample images from YakReID-103 and ATRW and ELPephants}. In each row, the first two images are of the same entity while the last image is of another. }
  \label{fig:reid}
\end{figure}

Telling apart different entities of the same animal species is a subtle task, as shown in  \cref{fig:reid}, one really needs to look at specific details of the distinguishing body part. For tigers, identification is usually based on body stripe patterns \cite{tiger_stripe}. For yaks, it is mostly their horns, sometimes fur and texture can help. For elephants, ears and tusks are important for identification. Matching the same body part between different entities is key to improving the Re-ID performance. 

\section{Ablation study.}
To understand which element of our model contributes to its overall performance we conduct an extensive ablation study,~\cref{tab:abla} on ATRW, to test the impact of each individual loss ($L_{DVE}$, $L_{ID}$, $L_{reID}$, $L_{LR}$) as well as the batch sampling strategy. The metric-based loss
$L_{reID}$ performs better than the classification loss $L_{ID}$ as the former learns more generalized feature representation. The use of orientation loss $L_{LR}$ in conjunction with $L_{reID}$ and $L_{ID}$ consistently boosts the performance, suggesting the importance of the pose information. $L_{DVE}$ loss and batch sampling, together, further improve the mAP score by $2.4$\%. This makes it evident that DVE loss helps to leverage animal part-specific information for better Re-ID. All results reported in this ablation study are without re-ranking.

\begin{table}[t]
\setlength{\tabcolsep}{5.2pt}
\begin{center}
\renewcommand{\arraystretch}{1}
\begin{tabular}{c c c c c| c c} \toprule
\multicolumn{5}{c}{Model Components} & \multicolumn{2}{c}{ATRW} \\ \cmidrule(lr{.75em}){6-7}
$L_{DVE}$ & $L_{ID}$ & $L_{ReID}$ & $L_{LR}$ & B.S. &\multicolumn{1}{c}{mAP} & \multicolumn{1}{c}{R@1(c)} \\ \midrule
&  \checkmark && & & 54.6 & 74.9 \\
&  &  \checkmark && & 56.3 & 72.0 \\
&  \checkmark  &\checkmark & & & 54.4 & 69.7 \\
&  \checkmark  & &\checkmark  & & 61.4 & 81.1 \\
&  &\checkmark   &\checkmark  &  & 60.5 & 76.6 \\
&  \checkmark  & \checkmark  &\checkmark  && 63.4  & 78.9 \\
 \checkmark &  \checkmark  & \checkmark  &\checkmark  && 63.4 & 78.3 \\
  &  \checkmark  & \checkmark  &\checkmark  &\checkmark & 63.2 & 80.6 \\
  \checkmark &  \checkmark  & \checkmark  &\checkmark  &\checkmark & 64.9 & 82.9 \\
\bottomrule
\end{tabular}
\end{center}
\caption{\textbf{Ablation study.} We study the influence of 5 different components of our approach: The impact of our 4 objective functions, and the batch sampling strategy (B.S.).
} 
\label{tab:abla}
\end{table}




\begin{table*}[ht]
  \centering
  \begin{tabular}{lccccccccc}
    \toprule
    & & \multicolumn{4}{c}{ATRW} & \multicolumn{2}{c}{YakReID-103} & \multicolumn{2}{c}{ELPephants} \\
    \cmidrule(lr{.75em}){4-6}\cmidrule(lr{.75em}){7-8}\cmidrule(lr{.75em}){9-10}
    Methods & Org. Re-ID Task & Pose GT  & mmAP & R@1(s) & R@1(c) & mAP & R@1 & mAP & R@1 \\
    \midrule
    & & & \multicolumn{7}{c}{Masked Backgrounds} \\
    \cmidrule(lr{.75em}){4-10}
    CLIP-Re-ID R-50 \cite{li2023clip} & Person  & \xmark & 54.4 & 82.6  & 71.4 & 48.2 & 85.6 & 13.1 & 25.5 \\
    CLIP-Re-ID ViT \cite{li2023clip} & Person  & \xmark & 56.5 & 86.3 & 72.0 & 49.8 & 82.2 & 12.7 & 25.3 \\
    \cmidrule(lr{.75em}){4-10}
    PPGNet R-50 \cite{liu_part-pose_2019}  & Animal  & \checkmark & 68.3 & 81.2 & 81.1 & - &- & - &-\\
    ResNet50 \cite{DBLP:journals/corr/HeZRS15}  & Animal  & \xmark & 65.9 & 91.1 & 83.4& 60.9 & 86.0 & 20.0 & 33.9 \\
    ViT \cite{dosovitskiy2020image} & Animal  & \xmark & 65.5 & 90.3  & 79.4 &\textbf{61.3}& 88.4 & 20.6 & 36.8  \\
    PGCFL \cite{liu_pose-guided_2019} & Animal   & \xmark & 66.9 & 90.8  & \textbf{86.3} & 55.8 & 82.7 & 18.5 & 33.4  \\
    Ours & Animal   & \xmark & \textbf{68.6} & \textbf{92.0} & 84.6 & 61.0 & \textbf{89.4} & \textbf{24.3} & \textbf{38.7}  \\
    \midrule
    & & & \multicolumn{7}{c}{Original Backgrounds} \\
    \cmidrule(lr{.75em}){4-10}
    CLIP-Re-ID R-50 \cite{li2023clip} & Person  & \xmark &  63.6 & 94.0  & 81.7 & 41.4   & 82.2 & 3.7  & 10.8 \\
    CLIP-Re-ID ViT \cite{li2023clip} & Person  & \xmark & 61.4 &  90.0 & 83.4  & 50.0 & 83.7  & 3.9  & 12.6  \\
    \cmidrule(lr{.75em}){4-10}
    PPGNet R-50 \cite{liu_part-pose_2019} & Animal   & \checkmark & 77.9& 99.4&90.8 &- &- & -& -\\
    PGCFL \cite{liu_pose-guided_2019} & Animal   & \xmark & 74.5 & \textbf{95.7}  & \textbf{90.3} & 64.3 & 91.8 & 10.7 & 24.7  \\
    PCN-RERP \cite{li2023automatic} & Animal & \xmark & - & - & - & \textbf{68.6}	& 91.8 & - & - \\
    Ours & Animal  & \xmark & \textbf{76.2} & \textbf{95.7} & 88.0 & 66.1 & \textbf{92.3} & \textbf{14.1} & \textbf{30.3}  \\       
    \bottomrule
  \end{tabular}

\caption{\textbf{Comparison to State-of-the-Art:} Our method is evaluated on three datasets, each representing a different species: ATRW (Tiger), YakReID-103 (Yak), and ELPephants (Elephants), originally proposed for various Re-ID tasks. Results from images with masked backgrounds are highlighted. Our model achieves top performance, surpassing existing baselines in mAP on ATRW and ELPephants, even outperforming PPGNet, which utilizes extra pose labels. When considering original images, our model outperforms PGCFL across all datasets and matches PPGNet on ATRW. CLIP-Re-ID, designed for person Re-ID, fails to generalize to animals due to high intra-class variations. PCN-RERP performs well on Yak dataset but lacks generalization to non-standing animal postures.}
\label{tab:sota}
\end{table*}

\paragraph{}

\section{Detailed Comparison to State-of-the-Art}
\paragraph{Baselines.} We compare our approach to PGCFL \cite{liu_pose-guided_2019}, PPGNet \cite{liu_part-pose_2019} the winner of the ATRW challenge, CLIP-Re-ID \cite{li2023clip} a recent person-reid method, and two other baselines: First a simple baseline using a ResNet50 backbone trained with $L_{base} = L_{ID} + L_{LR} +\lambda_{reID} \times  L_{reID}$. Then, a baseline using a more modern ViT \cite{dosovitskiy2020image} backbone is trained with $L_{base}$ as well. PPGNet is run for $160$ epochs and we followed the experimental settings from the original work. For the two training stages of CLIP-Re-ID, the ResNet-based model run for $120$ and $60$ epochs, and the ViT-based model run for $60$ and $80$ epochs. The rest of the experimental settings adhere to the CLIP-Re-ID original work. For all other methods, we use the same experimental setting previously described. PPGNet relies on groundtruth poses and therefore the evaluation is only shown on the ATRW dataset. Additionally, we compared with PCN-RERP~\cite{li2023automatic} who report their method on YakReID-103 dataset.

\paragraph{Results.} In \cref{tab:sota}, our approach outperforms prior state-of-the-art models on the ATRW dataset for mmAP and R@1(s). On the YakReID-103, it matches the ViT baseline in mAP, and outperforms all in terms of R@1. On the ELPephants dataset, it significantly outperforms all baselines with a margin of 5.8 mAP points w.r.t. PGCFL. CLIP-Re-ID baseline originally proposed for person re-id task fails to generalize on the animal dataset. This can be attributed to the fact of higher intra-identity variations occurring in animals than in persons, which CLIP-based models fail to capture. CLIP-based models are better known for their zero-shot inter-class/identity classification.  For completeness, we also provide results on the original benchmark without masking backgrounds. Here, overall metrics are higher but the ranking of the different baselines is similar. PCN-RERP~\cite{li2023automatic} has better overall mAP on Yak dataset but this approach cannot be generalized to dataset where animals are in the non-standing posture.

\section{Background removal criterion}
In the proposed protocol to remove the background, we aggregate the object masks generated by SAM if they satisfy a criterion  $h(\ve{m};S_{ISNet}(\ve{x}))\ge \Theta$. The function $h$ is manually tuned to better fit each dataset. We list the criteria used for each dataset here.

For ATRW we used:

\begin{equation}
    h(\ve{m};S_{ISNet}(\ve{x})) = \frac{|\ve{m} \cap S_{ISNet}(\ve{x}))|}{|\ve{m}|\cup |S_{ISNet}(\ve{x})|} \ge 0.3
\end{equation}

For ELPephants we used:

\begin{equation}
    h(\ve{m};S_{ISNet}(\ve{x})) =  \frac{|\ve{m} \cap S_{ISNet}(\ve{x}))|}{|\ve{m}|} \ge 0.5
\end{equation}

For YakReID-103 we directly used the output from SAM predictor, no filtering was applied.

\begin{figure}[h]
  \centering
  \includegraphics[width=0.75\linewidth]{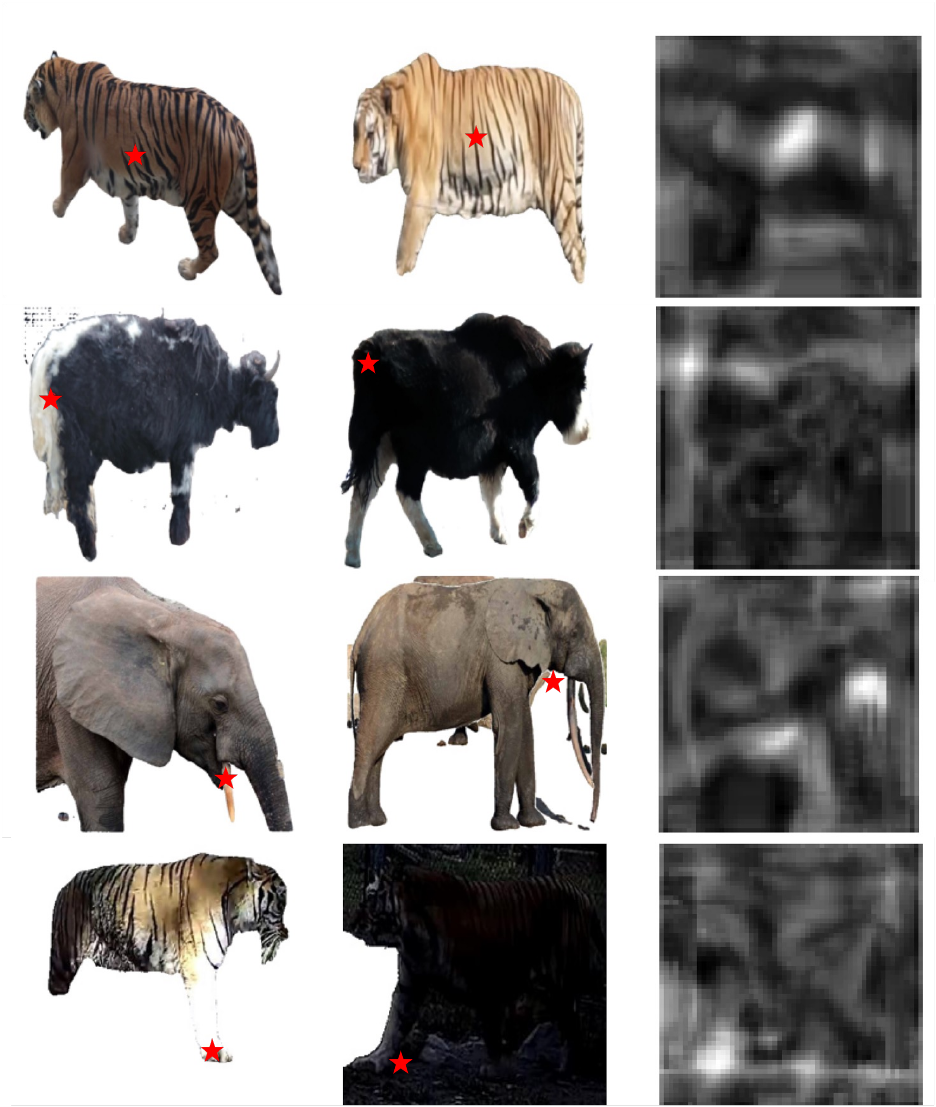}
  \caption{\textbf{Intra-Species visualization of SEResNet trained with DVE loss.} In each row, the red point from the left-most image is queried in the second image, and the red point in the middle image is its matching point, the right-most image is a heatmap of cosine similarities between the middle image and the red point in the left-most image. Interestingly, the last row show that even when images are of bad quality and of different views, the matching can be good. }
  \label{fig:dve_intra_viz}
\end{figure}
\begin{figure}[h]
  \centering
  \includegraphics[width=0.75\linewidth]{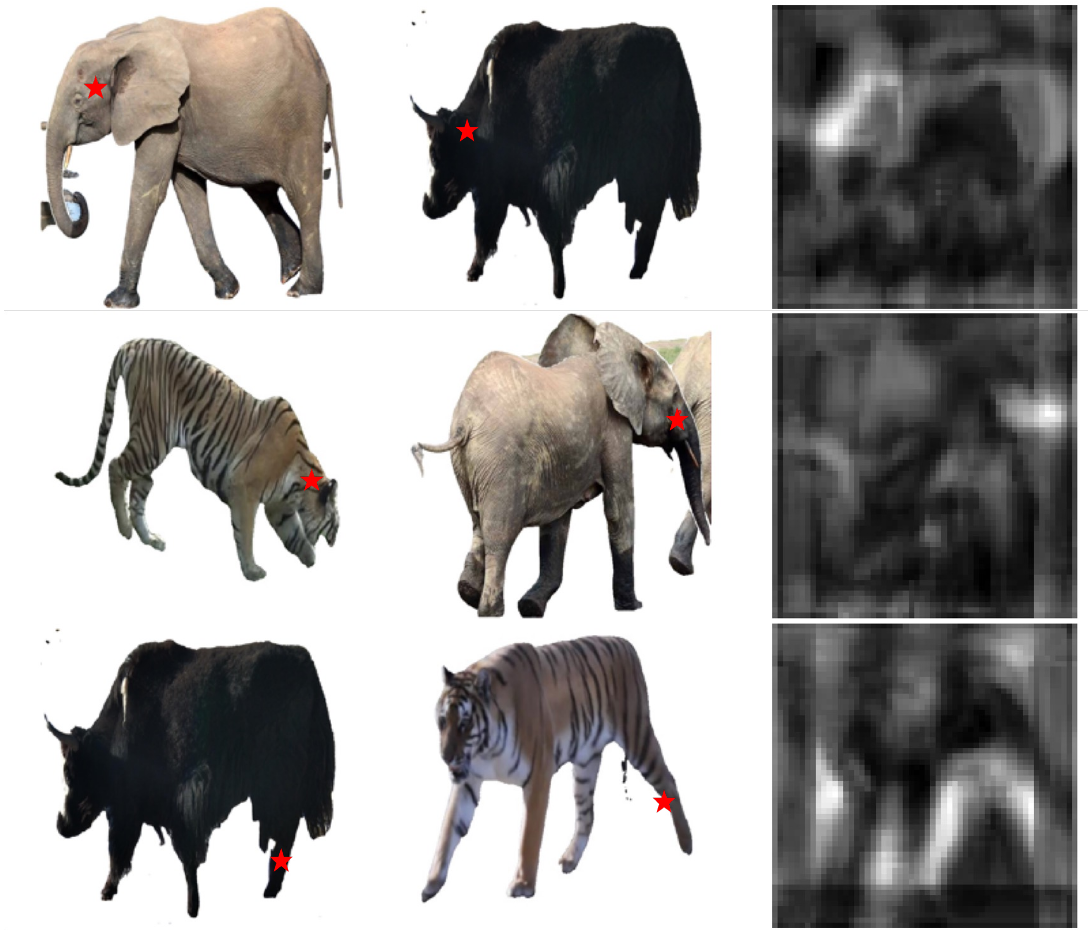}
  \caption{\textbf{Inter-Species visualization of SEResNet trained with DVE loss.} In each row, the red point from the left-most image is queried in the second image, and the red point in the middle image is its matching point, the right-most image is a heatmap of cosine similarities between the middle image and the red point in the left-most image. }
  \label{fig:dve_inter_viz}
\end{figure}
\begin{figure}[h]
  \centering
  \includegraphics[width=0.75\linewidth]{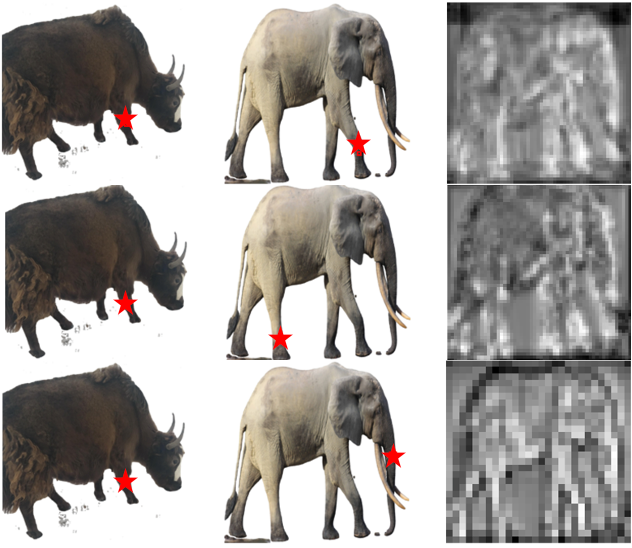}
  \caption{\textbf{Visualization of yak-elephant body part matching using features from the third layer of different models trained on tigers.} In each row, the red point from the left-most image is queried in the second image, and the red point in the middle image is its matching point, the right-most image is a heatmap of cosine similarities between the middle image and the red point in the left-most image. From top to bottom, each row correspond to our model, PGCFL and Resnet50 respectively.}
  \label{fig:transfer_l3}
\end{figure}

\section{Additional Qualitative Results}\label{sec:transferAddQuali}
We provide further results for transfer between species. In \cref{fig:dve_intra_viz} we show that the local descriptors learned by our approach also work for models trained on elephants or yaks, even in challenging scenarios with low-quality images (and failed background removal). In \cref{fig:dve_inter_viz} we provide additional results for the matching of local DVE features across species. Finally in \cref{fig:transfer_l3} we compare this transfer capability with two baselines: PGCFL and ResNet50. In both cases, they fail to match body parts, confirming that the matching capacity of our model is indeed coming from the use of $L_{DVE}$.

\section{Transfer Quantitative Evaluation}\label{sec:transferAddQuanti}
We provide a quantitative evaluation of transfer between species. Results can be found in \cref{tab:transfer}. While performance on transfer is lower than the supervised model, the model still manages to transfer some knowledge across species. Note that those performances are obtained on the masked background so the model can only rely on the animal's appearance.

\begin{table*}[ht]
  \centering
  \begin{tabular}{lccccccc}
    \toprule
    & \multicolumn{3}{c}{ATRW} & \multicolumn{2}{c}{YakReID-103} & \multicolumn{2}{c}{ElPephants} \\
    \cmidrule(lr{.75em}){2-4}\cmidrule(lr{.75em}){5-6}\cmidrule(lr{.75em}){7-8}
    Training Data     & mmAP & R@1(s) & R@1(m) & mAP & R@1 & mAP & R@1 \\
    \midrule
    ATRW & 68.6 & 92.0  & 84.6 & 35.9 & 73.1 & 6.7 & 17.6  \\
    YakReID-103 & 49.5 & 80.9  & 70.9 & 61.0 & 89.4  & 6.5 & 16.3  \\
    ElPephants & 47.2 & 80.9 & 66.3 & 34.5 & 75.0 & 24.3 & 38.7  \\     
    \bottomrule
  \end{tabular}

\caption{\textbf{Evaluation of inter-species transferability} We propose to evaluate the transferability of the proposed approach between three species: Tiger, Yak and Elephant. Each row of the table correspond to our model trained on a single species (Training data) and evaluated on the test set of the three species. While the performance on transfer are below than the fully supervised one, the proposed model is able to transfer meaningful representation between species.}
\label{tab:transfer}
\end{table*}

\section{Hyperparameters}
Our method relies on two main hyperparameters, $\lambda_{DVE}$ and $\lambda_{reID}$. For $\lambda_{DVE}$ we tested values in the range $[0,2]$ with $0.2$ increments and settled on the value $0.2$. For $\lambda_{reID}$ we tested values $1, 2, 5$ and settled for $2$ .



\section{Code}
The code is provided in the following link: \url{https://github.com/Chloe-Yu/Animal-Re-ID}. It contains all the scripts necessary to generate the masked background benchmarks and train and evaluate the proposed model.

\section{Data}
We provide the complete heading direction annotation for the ELPephant dataset in re{\_}mapped{\_}filtered{\_}anno{\_}elephant.csv together with our code.

\end{document}